# Hallucinated-IQA: No-Reference Image Quality Assessment via Adversarial Learning

Kwan-Yee Lin[1] and Guanxiang Wang[2]

[1]Department of Information Science, School of Mathematical Sciences, Peking University
[2]Department of Mathematics, School of Mathematical Sciences, Peking University

[1]linjunyi@pku.edu.cn  [2]gxwang@math.pku.edu.cn

## Abstract

*No-reference image quality assessment (NR-IQA) is a fundamental yet challenging task in low-level computer vision community. The difficulty is particularly pronounced for the limited information, for which the corresponding reference for comparison is typically absent. Although various feature extraction mechanisms have been leveraged from natural scene statistics to deep neural networks in previous methods, the performance bottleneck still exists.*

*In this work, we propose a hallucination-guided quality regression network to address the issue. We firstly generate a hallucinated reference constrained on the distorted image, to compensate the absence of the true reference. Then, we pair the information of hallucinated reference with the distorted image, and forward them to the regressor to learn the perceptual discrepancy with the guidance of an implicit ranking relationship within the generator, and therefore produce the precise quality prediction. To demonstrate the effectiveness of our approach, comprehensive experiments are evaluated on four popular image quality assessment benchmarks. Our method significantly outperforms all the previous state-of-the-art methods by large margins. The code and model are publicly available on the project page* https://kwanyeelin.github.io/projects/HIQA/HIQA.html.

## 1. Introduction

Image quality assessment (IQA) refers to the challenging task of automatically predicting the perceptual quality of a distorted image. IQA serves as a key component in the low-level computer vision community and has a wide range of applications [13, 26, 49].

IQA algorithms could be classified into three categories: full-reference IQA (FR-IQA) [50, 24, 19], reduced-reference IQA (RR-IQA) [11], and general purpose no-reference IQA (NR-IQA) [46, 17, 42, 51, 44, 20, 25]. Al-though FR-IQA and RR-IQA metrics have achieved remarkable results over the decades, the precondition of them that requiring a corresponding non-distorted reference image for comparison during quality predicting process makes these metrics infeasible in practical applications, since it is hard, even impossible in most cases, to obtain an ideal reference image. In contrast, NR-IQA, which takes only the distorted image to be assessed as input without any additional information, is more realistic and therefore receives substantial attention in recent years. However, the ill-posed definition makes it is highly challenging for NR-IQA to make a good image quality prediction.

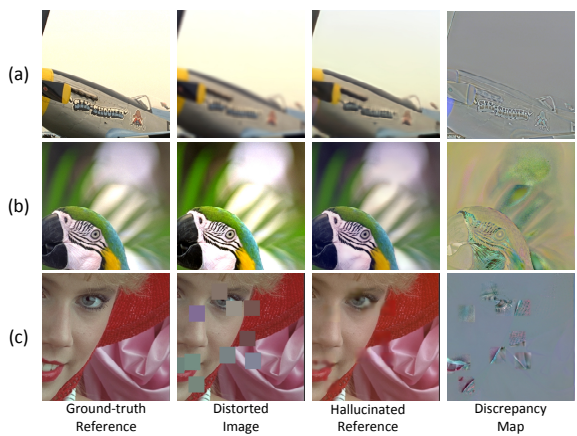

Figure 1: An illustration of our motivation. The first column is the Ground-truth Reference image which is undistorted. The second column shows several kinds of distortion that is easily happened. The third row demonstrates the hallucinated reference images which are generated by our approach. The fourth column is the discrepancy map which captures rich information that can be utilized to guide the learning of quality regression network to get high accuracy results.

The ill-posed nature of the underdetermined NR-IQA problem is particularly pronounced for the limited information, for which the form of distortion and the correspond-ing non-distorted reference image are typically absent. It is counter-intuitive, since human visual system (HVS) needs a reference to quantify the perceptual discrepancy by comparing the distorted image either directly with the original undistorted image or implicitly with a hallucinated scene in mind, as demonstrated in Figure. 1(a). The ill-posed definition becomes the most essential issue of NR-IQA task that leads to the performance bottleneck over the last decade.

Numerous efforts have been made to ease this problem by designing powerful feature representation models. Traditional methods commonly use manually designed statistic representations, and hence lack of diversity and flexibility for modeling the multiple complex distorted types[1] and large span of image contents (*e.g.*, human, animal, plant, cityscape, transportation, etc.) in NR-IQA. In recent years, the promising results of Deep Neural Networks (DNNs) in many computer vision tasks [8, 5, 43] encourage researchers exploring their formidable feature representation power to the NR-IQA task. Nevertheless, the extremely limited annotation samples in public datasets greatly limit the advantage of DNNs in NR-IQA task. To better leveraging the power of DNNs, previous works usually utilize various multi-task and data augmentation strategies with extra annotated ranking, proxy quality scores, or distortion information sophisticatedly, which are unavailable in practical NR-IQA applications, and hence lack of feasibility for modeling unknown multiple distortion types. Some works attempt to transfer general image feature representations from a pre-trained model on ImageNet [6] to quality prediction. While, less correlation and similarity between NR-IQA and image classification task reduce the effectiveness of transfer learning.

In this work, a Hallucination-Guided Quality Regression Network is proposed to simulate the behaviour of human visual system (HVS), which can make precise prediction by leveraging perceptual discrepancy information between the distorted image and hallucinated reference. As shown in Fig. 2, a high-resolution *scene hallucination* is firstly generated from the distorted image. Then, the discrepancy map which naturally encoding the difference between the distorted images and hallucinated reference can be obtained to guide the learning of regression network. With the strong and clear defined discrepancy information incorporated in, the ill-posed nature of NR-IQA can be dramatically overcome. Therefore, even with a common data augmentation, our approach could lead to better performance than all of the conventional sophisticated methods.

A straightforward way to generate the hallucinated reference is to leverage the state-of-the-art image super-resolution [23, 22, 40], blind deblur [33], or inpainting [34] methods to reconstruct images from the distorted ones. However, since an image could be distorted with multiple unknown distortions, which breaks the basic assumptions[2] of these related fields, it is impractical to utilize them to obtain a reconstructed image that qualified for the agent reference of NR-IQA task. To this end, a Quality-Aware Generative Network is proposed to generate hallucinated reference with a novel quality-aware perceptual term which is designed specifically for the NR-IQA task at hand.

While the Quality-Aware Generative Network is robust to most distortion types and levels, it is, however, still very challenging for a method that under the framework of DCNN to reconstruct high-frequency details with realistic texture, when the distorted image lacks structure information, as shown in Fig. 1(c). Since the result of hallucinated reference is crucial for final prediction, a bad hallucination will introduce large bias and therefore lead the regression results into sub-optimal values. We propose to tackle this problem with the two following mechanisms from low-level to high: (1) We introduce adversarial learning idea to hallucinated reference generation and quality prediction with a novel IQA-discriminator to, on the one hand, encourage the generated hallucinated scene perceptually hard to distinguish from the true reference images, and on the other hand, in a *low-level semantic*, constrain the influence of bad hallucinations to the quality regression network. (2) A novel *high-level semantic* fusion mechanism is introduced to further reduce the instability of the quality regression network caused by the hallucination model. It explores implicit ranking relationship within the hallucination network to as a guidance to help the regression network adjusting the image quality prediction in an adaptive manner. The quality-aware generative network, hallucination-guided quality regression network, and the iqa-discriminator can be jointly optimised in an end-to-end manner.

Our main contributions of this work are summarised into three folds:

- A novel Hallucination-Guided Quality Regression Network is proposed to incorporate the perceptual discrepancy information into network learning to overcome the ill-posed nature of NR-IQA and significantly improves the prediction precision and robustness.

- A Quality-Aware Generative Network together with a quality-aware perceptual loss is proposed, in which both texture feature similarity and quality feature similarity are taken into consideration in a complementary manner to help generating qualified hallucinated references.

---

[1] An image could be distorted by any stages in the whole process of its lifecycle from acquisition to storage, and therefore will undergo diverse distortions, like noise corruption, compression artifact, transmission errors, under-/over-exposure,etc. For more details, please refer to [35].

[2] For example, super-resolution methods usually assume the blur kernel or form is known.

- Since the result of hallucinated reference is crucial for final prediction, an IQA-Discriminator and an implicit ranking relationship fusion scheme are introduced to better guide the learning of generator and suppress the negative scene hallucination influence to quality regression in a low-level to high-level manner.

We evaluate the proposed method on four broadly used image quality assessment benchmarks including LIVE [41], CSIQ [21], TID2008 [36], and TID2013 [35]. Our approach shows the superior performance over all of the state-of-the-art NR-IQA methods by significant margins. Comprehensive ablation study further demonstrates the effectiveness of each component.

## 2. Related Work

**No-reference Image Quality Assessment.** In the literature of NR-IQA, besides classic methods ([31, 38, 29, 47]) and their improved versions ([51, 44, 27]), recently, significant progresses have been achieved by exploring DNNs for better feature representation [17, 18, 45, 20, 25, 48]. For example, Kang *et al*. [17] introduce a shallow ConvNet to model the quality prediction. This approach is refined to a multi-task CNN [18], where the network learns both distortion type and quality score simultaneously. Bianco *et al*. [2] use a pre-trained DCNN fine-tuned on an IQA dataset to extract features, then map them to IQA scores by an SVR model. Hui *et al*. [48] also propose to extract features by a pre-trained ResNet [14]. Instead of learning IQA scores directly, they fine tune the network to learn a probabilistic representation of distorted images. According to the distortion types and levels in particular datasets, Liu *et al*. [25] synthesize masses of ranked images to train a Siamese network to learn the rankings for NR-IQA. Liang *et al*. [24] propose to use non-aligned similar scene as a reference. Kim and Lee [20] apply state-of-the-art FR-IQA methods to generate proxy scores on patches as the ground truth to pre-train the model and then fine-tune to NR-IQA.

In this work, we propose a unique approach to address the ill-posed problem by compensating the absent reference information without any extra data annotation or prior knowledge, which therefore increases the flexibility and feasibility than other methods.

**Generative Adversarial Network.** GANs [12] and various variants [37, 1, 28] flourish in generating natural images such as human faces [3] and indoor scenes [7]. However, generating high-resolution images (*e.g*.,256×256) will lead GANs to training instability and sometimes nonsensical outputs, which has been proven in [16]. Since our ultimate goal is NR-IQA, and the performance of quality regression network is closely related to the output of the generator, instead of applying original discriminator, we tailor the adversarial learning scheme for image quality assessment by introducing an effective iqa-discriminative network.

## 3. Our Approach

In this section, we introduce our approach for NR-IQA. An overview of our framework is illustrated in Fig. 2. The model consists of three parts, *i.e*., the quality-aware generative network $G$, the iqa-discriminative network $D$ and the hallucination-guided quality regression network $R$. The generative network produces hallucinated reference as the compensatory information for the distorted images. The discriminative network is trained with $G$ in the adversarial manner to help $G$ producing more qualified results and constrain negative effects of bad ones to $R$. We define the objective discrepancy (*i.e*., the pixel-wise differences) between a distorted image and the corresponding scene hallucination as the *discrepancy map*[3]. The quality regression network takes the distorted images and corresponding discrepancy maps as inputs, with the guidance of implicit ranking relationships in $G$, to exploit the perceptual discrepancy and produce the predicted quality scores as outputs.

### 3.1. Quality-Aware Generative Network

As we mentioned in the previous sections, the function of hallucinated reference for the distorted image is to compensate the absence of true reference image, and the less gap between hallucination and true reference, the more precise the quality regression network will perform. Therefore, the aim of $G$ is to generate a high-resolution hallucinate image $I_{sh}$ conditioned on the distorted image $I_d$. Toward this end, we adopt a stacked hourglass [32] as the baseline of the generative network.

A straightforward way for learning the generating function $G(I_d)$ is to enforce the output of the generator both pixel-wise and perception-wise close to the true reference. Therefore, given a set of distorted images $\{I_d^i, i = 1, 2, \ldots, N\}$, and corresponding true reference images $\{I_r^i, i = 1, 2, \ldots, N\}$, we solve

$$\hat{\theta} = \arg\min_{\theta} \frac{1}{N} \sum_{i=1}^{N}(l_p(G_\theta(I_d^i), I_r^i) + l_s(G_\theta(I_d^i), I_r^i)), \quad (1)$$

where $l_p$ penalizes the pixel-wise differences between the output and the ground truth with pixel-level error measurements, such as MSE, to generate holistic content; and $l_s$ penalizes the perception-wise differences to achieve sharper local results. We adopt a feature space loss term [9] as the perception constraint, which is defined as

$$l_s(G_\theta(I_d^i), I_r^i) = \|\phi(G_\theta(I_d^i)) - \phi(I_r^i)\|^2, \quad (2)$$

---
[3]This is different from the concept of *error map*, which is used in FR-IQA to represent pixel-wise error between the distorted image and true reference.

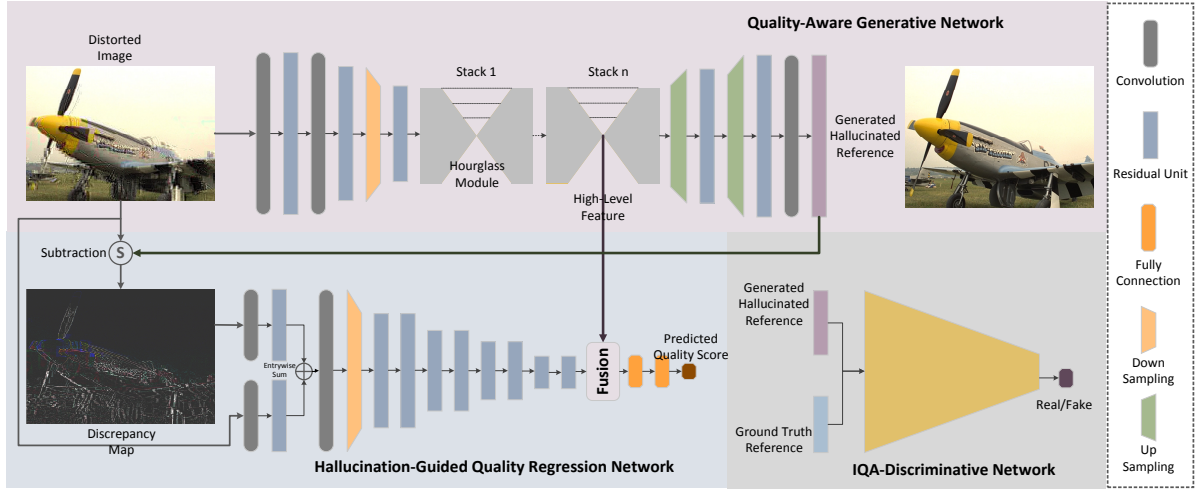

Figure 2: An illustration of our proposed Hallucinated-IQA framework. It consists of three strongly related subnets. (a) Quality-Aware Generative Network is used to generate hallucinated reference images. In order to get high resolution hallucinated images, a quality-aware loss is introduced to the learning process. (b) Hallucination-Guided Quality Regression Network is in a position to incorporate the discrepancy information between the hallucinated image and distorted encoded in the discrepancy map. The incorporated discrepancy information together with high-level semantic fusion from the generative network can supply the regression network with rich information and greatly guide the network learning. (3) Since the results of the hallucinated image are crucial for the final prediction, IQA-Discriminator is proposed to further refine the hallucinated image.

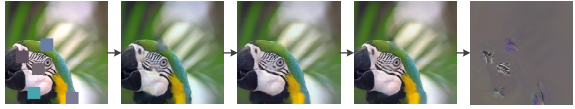

Figure 3: An illustration of the effectiveness of quality-aware loss and IQA-GAN. With the quality-aware loss and IQA-GAN scheme adding, the hallucinated images are improved to be more and more clear and plausible. The last column shows the discrepancy map got from our model, which can be seen to well capture the type and location information of the distortion. The map is demonstrated to be very helpful for our IQA task.

where $\phi(\cdot)$ represents a feature transformation. Intuitively, pre-trained network like VGG-19 could be utilized to calculate the perception term. This is reasonable in most cases by the fact that the VGG-19 is trained for semantic classification, and the features of its intermediate layers are therefore invariant to the noise of input [4, 10]. Consequently, these layers provide structure and texture information to the generator for inferring more accurate results. However, the invariance property will also lead to the perception term ignoring the hard cases where the output of the generator still contains a certain degree distortion information, as demonstrated in Fig. 3. To ease this problem, we propose a quality-aware perceptual loss, which incorporates the features of the deep regression network $R$ dynamically. The loss function in equation (2) becomes

$$l_s(G_\theta(I_d^i), I_r^i) = \lambda_1 l_v(G_\theta(I_d^i), I_r^i) + \lambda_2 l_q(G_\theta(I_d^i), I_r^i), \quad (3)$$

where

$$l_v = \sum_{c_v=1}^{C_v} \frac{1}{W_j H_j} \sum_{x=1}^{W_j} \sum_{y=1}^{H_j} \|\phi_j(G_\theta(I_d^i))_{x,y} - \phi_j(I_r^i)_{x,y}\|^2, \quad (4)$$

and

$$l_q = \sum_{c_q=1}^{C_q} \frac{1}{W_k H_k} \sum_{x=1}^{W_k} \sum_{y=1}^{H_k} \|\pi_k(G_\theta(I_d^i))_{x,y} - \pi_k(I_r^i)_{x,y}\|^2, \quad (5)$$

where $\phi_j(\cdot)$ denotes the feature map at the $j$-th layer of VGG-19, $\pi_k(\cdot)$ denotes the feature map at the $k$-th layer of $R$; $W$ and $H$ represent the dimensions of the feature map, $C$ represents the number of feature maps at a particular layer. Since the vgg-19 network and $R$ are trained for different tasks, the representation of kernels within the two networks also toward to preserve different information. The activations from the layers of a pre-trained [4] NR-IQA regression network capture the distortion information of the input, which ensures the quality similarity measurement between the output of $G$ and the ground truth. The activations from the layers of the VGG-19 network ensure the semantic similarity measurement. Base on respective representing capabilities of the two networks, incorporating both $l_v$ and $l_q$ losses to the perception term could complement each other and therefore help the generator producing better results jointly.

---

[4] It should be noted that, the pre-trained quality regression model refers to the one that trained from scratch with IQA dataset.

## 3.2. IQA-Discriminative Network

To ensure the generator producing high perceptual outputs with realistic high-frequency details, especially for the samples that seriously lack structure and texture information due to the distortion type (*e.g.*, local block-wise distortions of different intensity, transmission errors), or the distortion level, the adversarial learning mechanism is introduced to our work.

The original manner of adversarial learning is to train $G$ to generate images to fool $D$, and $D$ is in contrast trained to distinguish *fake* reference images $I_{sh}$ from *real* reference images $I_r$. However, since GANs are limited to the resolution of the generator, and the distorted images forwarded to a quality network are usually of large size to maintain sufficient contextual information, directly providing $I_{sh}$ as fake images to the discriminator will introduce instability to optimization procedure and sometimes leads to nonsensical results. More importantly, our ultimate goal is improving the performance of the deep regression network $R$. Even when $G$ fails to generate high-resolution hallucination images, the predicted scores of $R$ should still be a reasonable value. Thus, the influence of bad hallucination images to $R$ should be suppressed. Thus, we propose a IQA-Discriminator (*i.e.*, $D$) to ease above problems by discriminating the fake samples from the real samples according to their positive or negative influence to $R$. If $G$ generates a hallucinated reference could help improving the precision of $R$, then this hallucination is defined as *real* sample to $D$, otherwise the hallucination is a *fake* sample. This could be formulated as

$$\max_{\omega} \mathbb{E}[\log D_\omega(\mathbf{I}_r)] + \mathbb{E}[\log(1 - |D_\omega(G_\theta(\mathbf{I}_d)) - \mathbf{d}_{fake}|)], \quad (6)$$

where $\mathbf{d}_{fake}$ denotes the ground truth influence label with the definition

$$\mathbf{d}_{fake}^i = \begin{cases} 1 & \text{if } \|R(I_d^i, I_{sh}^i) - s^i\|_F < \epsilon \\ 0 & \text{if } \|R(I_d^i, I_{sh}^i) - s^i\|_F \geq \epsilon \end{cases}, \quad (7)$$

where $s^i$ is the ground truth quality score of $I_d^i$, $\epsilon$ denotes the threshold parameter. The general idea behind this formulation is that it leverages the property of quality regression loss, where the loss is an explicit index that directly reflects the impact of $G$ on $R$, to enforce $D$ only penalizing samples with negative influence. Therefore, it could also be regarded as a relaxation strategy to stabilize the adversarial learning process.

Thus, $G$ is eventually optimised to fool the discriminator $D$ by generating qualified hallucinated scene that is beneficial for $R$. The adversarial loss of $G$ is formulated as

$$\mathcal{L}_{adv} = \mathbb{E}[\log(1 - D_\omega(G_\theta(I_d)))], \quad (8)$$

and the overall loss function of $G$ for all training samples is given by

$$\mathcal{L}_G = \mu_1 \mathcal{L}_p + \mu_2 \mathcal{L}_s + \mu_3 \mathcal{L}_{adv}, \quad (9)$$

where $\mu_1$, $\mu_2$ and $\mu_3$ represent the parameters that trade off the three loss components.

## 3.3. Hallucination-Guided Quality Regression Network

Given the hallucinated scene generated by $G$, we are able to provide agent references to the quality regression network to compensate the absence of true reference information. In order to incorporate the hallucinated reference information effectively, the concept of *discrepancy map* is introduced to the work. To further stabilizing the optimization procedure of $R$, a high-level semantic fusion scheme is proposed.

**Discrepancy Map.** Given a set of distorted images to be assessed, previous CNN-based NR-IQA methods learn a mapping function $\mathcal{R}(I_d)$ to predict the quality scores. On the contrary, we consider the distorted images and their discrepancy maps as pairs $\{I_d^i, I_{map}^i\}_{i=1}^N$ to train a deep regression network by solving

$$\hat{\gamma} = \arg\min_{\gamma} \frac{1}{N} \sum_{i=1}^N l_r(\mathcal{R}(I_d^i, I_{map}^i), s^i), \quad (10)$$

where $I_{map} = |I_d - G_{\hat{\theta}}(I_d)|$, denotes the discrepancy map. The formulation shows the discrepancy map could virtually be regarded as a prior information to tell the network what the distortion looks like.

It is interesting that, so far, the holistic mechanism functions in a reinforced way, during training stage of $R$, $G$ is used to produce auxiliary hallucinated references, while during the training stage of $G$, $R$ is in reverse introduced to help generating better hallucinations. In essence, $G$ and $R$ are mutually correlated and thus can reinforce each other.

**High-level Semantic Fusion.** As we mentioned in previous sections, the precision of $R$ is greatly depended on the eligibility of the hallucinated scene. To be specific, a qualified hallucination as the agent reference could help $R$ exploring correct perceptual discrepancy of the distorted image, while the unqualified one will conversely introduce large bias to $R$ by improperly narrowing the distortion information. Hence, a constrained scheme is needed to stabilize the quality regression process.

Assume $G$ has been trained, the feature maps after the $m$-th residual block in encoder part of its $n$-th stack are considered as $\{\mathcal{H}_{mn}^{c_{mn}}(I_d)\}_{c_{mn}=1}^{C_{mn}}$. We fuse the ones after the last encoder residual block of second stack with the feature maps after the last block of $R$, then we have the fusion term:

$$\mathcal{F} = f(\mathcal{H}_{5,2}(I_d)) \otimes (\mathcal{R}_1(I_d, I_{map})) \quad (11)$$

where $f$ is a linear projection to ensure the dimensions of $\mathcal{H}$ and $\mathcal{R}_1$ are equal, $\mathcal{R}_1$ denotes the feature extraction before the fully connected layers ($\mathcal{R}_2$) of $R$, and $\otimes$ denotes

the concatenation operation. Thus, the loss of $R$ could be formulated as:

$$\mathcal{L}_R = \frac{1}{T}\sum_{t=1}^{T}\|\mathcal{R}_2(f(\mathcal{H}_{5,2}(I_d)) \otimes (\mathcal{R}_1(I_d, I_{map}))) - s^t\|_{\ell 1} \quad (12)$$

The form of the loss $\mathcal{L}_R$ allows the high-level semantic information of $G$ participating in the optimization procedure of $R$. As we discussed in the introduction, the fusion term $\mathcal{F}$ explores implicit ranking relationship[5] within $G$ to as a guidance to help $R$ adjusting the quality prediction in an adaptive manner. Specifically, if $G$ is optimal, the solvers may simply drive the weights of neurons in $\mathcal{R}_2$ that connecting with $f$ toward zero to approach identity mappings. Otherwise, the eligibility of the hallucinated scene is materially a reflection of the quality of the input distorted images that could be leveraged to as a guidance to correct the prediction, and therefore improving the precision of $R$ in a *high-level semantic* manner. Meanwhile, the iqa-discriminator could be regarded as a *low-level semantic* scheme to $R$, since it encourages $G$ to generate useful hallucination input to $R$. Therefore, our model has schemes in multiple semantic levels to stabilize the quality regression process.

### 3.4. Training Strategy

Since all of the operations in $G$ and $R$ are differentiable, these two sub-networks can be trained in an end-to-end manner. To better optimize the generation and quality regression in a mutually reinforced way, we take an alternative training strategy in practice. Please refer to the supplementary, where Algorithm 1 demonstrates the whole training processing of our approach as the pseudo codes.

### 3.5. Weakly-Supervised Quality Assessment

In this section, we discuss some extensions to further uncover the potential of our framework.

To advance the development of IQA task, various benchmarks have been released in these years. However, a significant issue follows as well. As shown in Table 1, there are huge gaps of distorted quality definition, types and levels among the datasets. While NR-IQA models are commonly trained on one specific dataset, these gaps will easily lead the models to suffer over-fitting problem and lack of generalization ability. Learning from cross-datasets is an alternative way to ease the problem. Previous methods usually transfer the definition of quality scores by non-linear mappings learned from the distributions of datasets, which may introduce bias to the models.

---
[5]$G$ serves as not only a generator, but also an encoder-decoder mechanism. Thus, the difference-information between images distorted in different degree is encoded compactly in the end of the encoder part. We refer to this "difference-information" as "implicit ranking relationship" of distorted images in this work.

In contrast, as a by-product of our work, the hallucinated scene could be regarded as a universal medium among different datasets to help the training process of a particular one without losing precision, since the hallucination is only constrained on distorted image and serves as the fundamental agent reference information of image quality. Meanwhile, the detachable training process of our framework provides an alternative that the $R$ in the stages of training $G$ and final quality regression model could be different. Based on the above, as long as a hallucination generator is trained either on one specific dataset with multiple complex distortions or on multi-datasets in once, it can be used to help the training process of any other datasets as a *plug-and-play* module in a weakly-supervised manner. Moreover, the module could also be used as a data augmentation or initialization mechanism without any extra annotation or artificial prior knowledge. We evaluate above discussion in Sec.4.1.

### 3.6. Implementation Details

All the training samples are $256 \times 256$ pixel patches that randomly sampled from the original images. Then a common data augmentation is performed with random rotation ($\pm 20°$) and flip. We train our models with Caffe [15] on the Titan X GPUs with a mini-batch size of 32 and all of them are trained from *scratch*. The stochastic gradient descent (SGD) is used to optimise the networks with an initial learning rate of $10^{-5}$ for the generation network and $10^{-2}$ for the regression network, and dropped by a factor of $0.1$ every 20K iterations. The weight decay is 0.0005, and the momentum is 0.9. During testing, we extract overlapped image patches at a fixed stride from each testing image, and simply average all predicted scores as the final whole-image quality score.

| Databases | # of Ref.Images | # of Dist.Images | # of Dist.Types | Score Type | Score Range |
|---|---|---|---|---|---|
| LIVE | 29 | 779 | 5 | DMOS | [1,100] |
| CSIQ | 30 | 866 | 6 | DMOS | [0,1] |
| TID2008 | 25 | 1700 | 17 | MOS | [0,9] |
| TID2013 | 25 | 3000 | 24 | MOS | [0,9] |

Table 1: Summary of the databases evaluated in the experiments.

## 4. Experiments

**Datasets.** We perform experiments on four widely used benchmark datasets LIVE [41], CSIQ [21], TID2008 [36], and TID2013 [35]. The detailed information are summarized in Table 1.

**Evaluation Metrics.** Following most previous works, two evaluation criteria are adopted in our paper: the Spearman's Rank Order Correlation Coefficient (SROCC) and the Linear Correlation Coefficient (LCC). SROCC is a measure of the monotonic relationship between the ground-truth and model prediction. LCC is a measure of the linear correlation between the ground-truth and model prediction. The detailed definitions are formulated in the supplementary material.

| Methods | # 1 | # 2 | # 3 | # 4 | # 5 | # 6 | # 7 | # 8 | # 9 | # 10 | # 11 | # 12 | # 13 |
|---|---|---|---|---|---|---|---|---|---|---|---|---|---|
| BLIINDS-II [39] | 0.714 | 0.728 | 0.825 | 0.358 | 0.852 | 0.664 | 0.780 | 0.852 | 0.754 | 0.808 | 0.862 | 0.251 | 0.755 |
| CORNIA-10K [47] | 0.341 | -0.196 | 0.689 | 0.184 | 0.607 | -0.014 | 0.673 | 0.896 | 0.787 | 0.875 | 0.911 | 0.310 | 0.625 |
| HOSA [44] | 0.853 | 0.625 | 0.782 | 0.368 | 0.905 | 0.775 | 0.810 | 0.892 | 0.870 | 0.893 | **0.932** | **0.747** | 0.701 |
| RankIQA [25] | 0.667 | 0.620 | 0.821 | 0.365 | 0.760 | 0.736 | 0.783 | 0.809 | 0.767 | 0.866 | 0.878 | 0.704 | **0.810** |
| **Ours** | **0.923** | **0.880** | **0.945** | **0.673** | **0.955** | **0.810** | **0.855** | 0.832 | **0.957** | **0.914** | 0.624 | 0.460 | 0.782 |
| **Ours+Oracle** | 0.952 | 0.890 | 0.976 | 0.831 | 0.931 | 0.773 | 0.898 | 0.812 | 0.910 | 0.929 | 0.735 | 0.638 | 0.739 |
| Methods | # 14 | # 15 | # 16 | # 17 | # 18 | # 19 | # 20 | # 21 | # 22 | # 23 | # 24 | ALL | |
| BLIINDS-II [39] | 0.081 | 0.371 | 0.159 | -0.082 | 0.109 | 0.699 | 0.222 | 0.451 | 0.815 | 0.568 | 0.856 | 0.550 | |
| CORNIA-10K [47] | 0.161 | 0.096 | 0.008 | 0.423 | -0.055 | 0.259 | 0.606 | 0.555 | 0.592 | 0.759 | 0.903 | 0.651 | |
| HOSA [44] | 0.199 | 0.327 | 0.233 | 0.294 | 0.119 | 0.782 | 0.532 | 0.835 | 0.855 | **0.801** | **0.905** | 0.728 | |
| RankIQA [25] | 0.512 | **0.622** | 0.268 | **0.613** | **0.662** | 0.619 | **0.644** | 0.800 | 0.779 | 0.629 | 0.859 | 0.780 | |
| **Ours** | **0.664** | 0.122 | 0.182 | 0.376 | 0.156 | **0.850** | 0.614 | **0.852** | **0.911** | 0.381 | 0.616 | **0.879** | |
| **Ours+Oracle** | 0.834 | 0.457 | 0.823 | 0.850 | 0.539 | 0.893 | 0.695 | 0.859 | 0.910 | 0.655 | 0.712 | 0.935 | |

Table 2: Performance evaluation (SROCC) on the entire TID2013 database.

## 4.1. Comparisons with the state-of-the-arts

To validate our approach, we conduct extensive evaluations, where ten state-of-the-art NR-IQA methods are compared. We follow the experimental protocol used in three most recent algorithms (*i.e.*, HOSA [44], BIECON [20], and RankIQA [25]), where the reference images are randomly divided into two subsets with 80% for training and 20% for testing, and the corresponding distorted images are divided in the same way to ensure there is no overlap image content between the two sets. All the experiments are under ten times random train-test splitting operation, and the median SROCC and LCC values are reported as final statistics.

**Single dataset evaluations.** We first analyze the experiment results on TID2013. The SROCC for our approach and compared state-of-the-arts on entire TID2013 dataset are reported in Table 2. Our method significantly outperforms previous methods by a large margin. We achieve a 13% relative improvement over the most state-of-the-art method RankIQA on entire dataset with all the distortion types under consideration at once. For individual distortions, due to the normalization operation in the network, the performances on a small number of types like intensity shift and change of colour saturation are lower than some methods. While we generally achieve the highest accuracies on most distortion types (over 60% subsets). Specifically, the significant improvements on distortion types like #4(masked noise) and #14(non-eccentricity pattern noise) quantitatively demonstrate the effectiveness of our hallucinated reference compensation mechanism, and improvements on types such as #9 (Image denoising) and #22 (Multiplicative Gaussian noise) verify the capacity of our $G$ component as a single model that hallucinates images under multiple distortions effectively.

Table 3 shows the performance evaluation on the entire LIVE database. Our method outperforms all of the state-of-the-art methods for both SROCC and LCC evaluations. Among the methods compared in the experiments, the most state-of-the-art three methods explore different strategies to better leverage the power of DNNs and achieve promising results, where BIECON uses FR-IQA methods to generated proxy quality scores, RankIQA synthesizes masses of ranked images to train the network, and PQR takes advantage from a pre-trained Res-50 network. Our method achieves 2% improvements than BIECON, 2% SROCC and 1% LCC improvements than PQR, and 0.1% slightly improvements than RankIQA with training from scratch. These observation demonstrate that our mechanisms increase the model capacity effectively from a new perspective.

As for TID2008 dataset, our approach also achieves highest performances compared with all of the state-of-the-arts. We also reach best performances on CSIQ dataset. For space saving, the detail results and discussion of this two dataset are shown in the supplementary material, please refer to it.

We also list the results of using ground-truth reference on above experiments as the *theoretical bounds*, which are referred to "ours+oracle", to further verify the effectiveness and potential of proposed hallucinated references to NR-IQA. The oracle outperforms all the methods in all datasets by large margins. These results demonstrate the effectiveness of hallucinated information and show great potential performance gain if the hallucinated information could be well generated.

**Cross-dataset evaluations.** Here, we perform two types of cross-dataset evaluations to further verify some merits of our approach. Table 4 shows the results of cross-dataset test where the models are trained by the LIVE dataset, and tested on the TID2008 dataset. We follow the common experiment setting to test the results on the subsets of TID2008, where four distortion types (*i.e.*, JPEG, JPEG2K, WN, and BLUR) are included, and a logistic regression is applied to match the predicted DMOS to MOS value. The promising results demonstrate the generalization ability of our approach.

To evaluate the by-product of our work where the model could be leveraged in a weakly-supervised manner to han-dle cross-dataset quality assessment, we train the generator on different datasets, and use LIVE dataset to train the regression network. Table 5 reports the results. The "L" as the plain of the experiment represents the hallucination generator is trained on the training set of LIVE. The "T08" represents training the generator on TID2008, and "T08+T13" is the version that training on both the TID2008 and TID2013. It can be clearly observed that with more IQA datasets aggregated in the generator, the regression network reaches higher SROCC and LCC performances to approximate the oracle.

| SROCC | JP2K | JPEG | WN | BLUR | FF | ALL |
|---|---|---|---|---|---|---|
| BRISQUE [30] | 0.914 | 0.965 | 0.979 | 0.951 | 0.877 | 0.940 |
| CORNIA [47] | 0.943 | 0.955 | 0.976 | 0.969 | 0.906 | 0.942 |
| CNN [17] | 0.952 | 0.977 | 0.978 | 0.962 | 0.908 | 0.956 |
| SOM [51] | 0.947 | 0.952 | 0.984 | 0.976 | 0.937 | 0.964 |
| BIECON [20] | 0.952 | 0.974 | 0.980 | 0.956 | 0.923 | 0.961 |
| RankIQA [25] | 0.970 | 0.978 | 0.991 | 0.988 | 0.954 | 0.981 |
| PQR [48] | - | - | - | - | - | 0.965 |
| **Ours** | 0.983 | 0.961 | 0.984 | 0.983 | 0.989 | **0.982** |
| **Ours+Oracle** | 0.978 | 0.960 | 0.993 | 0.988 | 0.968 | 0.983 |
| LCC | JP2K | JPEG | WN | BLUR | FF | ALL |
| BRISQUE [30] | 0.923 | 0.973 | 0.985 | 0.951 | 0.903 | 0.942 |
| CORNIA [47] | 0.951 | 0.965 | 0.987 | 0.968 | 0.917 | 0.935 |
| CNN [17] | 0.953 | 0.981 | 0.984 | 0.953 | 0.933 | 0.953 |
| SOM [51] | 0.952 | 0.961 | 0.991 | 0.974 | 0.954 | 0.962 |
| BIECON [20] | 0.965 | 0.987 | 0.970 | 0.945 | 0.931 | 0.962 |
| RankIQA [25] | 0.975 | 0.986 | 0.994 | 0.988 | 0.960 | 0.982 |
| PQR [48] | - | - | - | - | - | 0.971 |
| **Ours** | 0.977 | 0.984 | 0.993 | 0.990 | 0.960 | **0.982** |
| **Ours+Oracle** | 0.989 | 0.985 | 0.997 | 0.992 | 0.988 | 0.989 |

Table 3: Performance evaluation (both SROCC and LCC) on the entire LIVE database.

| | CORNIA [47] | CNN [17] | SOM [51] | Ours | Ours+Oracle |
|---|---|---|---|---|---|
| SROCC | 0.892 | 0.920 | 0.923 | 0.934 | 0.939 |
| LCC | 0.880 | 0.903 | 0.899 | 0.917 | 0.920 |

Table 4: Cross-dataset evaluation (SROCC). The models are trained on the LIVE database and tested on the subset of TID2008.

| | L | T08 | T08+T13 | Ours+Oracle |
|---|---|---|---|---|
| SROCC | 0.982 | 0.982 | 0.983 | 0.983 |
| LCC | 0.982 | 0.985 | 0.988 | 0.989 |

Table 5: SROCC and LCC results of models on the LIVE database with training generator on different datasets.

### 4.2. Ablation study

To investigate the efficacy of the key components of our model, we conduct ablation experiments on the TID2008 dataset. The overall results are shown in Figure 4. We use a modified Res-18 network with only distorted images as inputs to be our baseline model and analyze each proposed component based on the baseline network (BL), by comparing both SROCC and LCC results.

**Hallucinated reference compensation.** We first evaluate the hallucinated reference compensation mechanism.

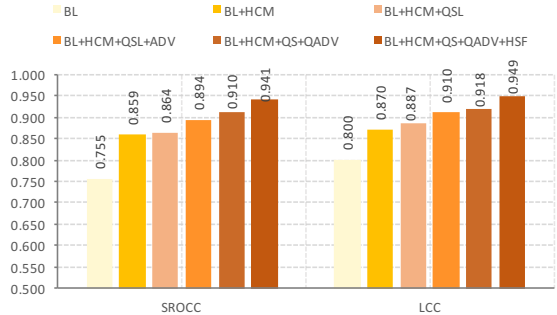

Figure 4: Ablation results on the entire TID2008 dataset.

By adding a holistic hallucination model to provide hallucinated references pairing with distorted images as the inputs to res-18 network ("BL+HCM"), we get a 0.859 SROCC value and a 0.870 PLCC value, which up to 14% and 8% improvement compared to the baseline model, respectively.

**Quality-aware perceptual loss.** By adding the feature matching loss w.r.t. quality similarity at the training process of the hallucination model("BL+HCM+QPL"), our model obtains a further 0.5% improvement on SROCC and 2% on LCC.

**Adversarial learning.** To explore the effect of proposed IQA-Discriminative network for quality assessment, we further compare the models with adversarial learning mechanism under original definition ("BL+HCM+QPL+ADV") and our definition("BL+HCM+QPL+QADV"). Adding original adversarial learning mechanism leads to a 3% improvement on SROCC and 3% on LCC, while our method obtains further 2% and about 1% improvements on SROCC and LCC, respectively.

**Multi-level semantic fusion.** We also show the improvements brought by the multi-level semantic fusion mechanism. We fuse the feature maps of the generator from stack two with the ones of same size in quality regression network, and obtain the highest 0.941 SROCC value and 0.949 LCC value.

## 5. Conclusion

In this paper, we propose to solve the ill-posed nature of NR-IQA from a new perspective. We introduce a hallucination-guided quality regression network to capture the perceptual discrepancy between the distorted images and the hallucinated images, and therefore predict precise perceptual quality result. We generate the hallucinations by a novel quality-aware generation network with the help of a specially designed iqa-discriminator under the adversarial learning manner. The proposed network does not require any extra annotations or artificial prior knowledge for training and can be trained end-to-end. Extensive experiments demonstrate the superior performance on NR-IQA task.